\documentclass[10pt,twocolumn,letterpaper]{article}

\usepackage[pagenumbers]{cvpr} 

\usepackage[dvipsnames]{xcolor}

\usepackage{tabularx}
\usepackage{boldline}

\definecolor{cvprblue}{rgb}{0.21,0.49,0.74}
\usepackage[pagebackref,breaklinks,colorlinks,citecolor=cvprblue]{hyperref}

\def\paperID{1439} 
\def\confName{CVPR}
\def\confYear{2024}

\title{Egocentric Whole-Body Motion Capture with FisheyeViT and Diffusion-Based Motion Refinement}

\author{Jian Wang\textsuperscript{1,2}~~~~~~ Zhe Cao\textsuperscript{3}~~~~~~Diogo Luvizon\textsuperscript{1,2}~~~~~~Lingjie Liu\textsuperscript{4}\\
Kripasindhu Sarkar\textsuperscript{3}~~~~~~Danhang Tang\textsuperscript{3}~~~~~~Thabo Beeler\textsuperscript{3}~~~~~~Christian Theobalt\textsuperscript{1,2}\\
	\textsuperscript{1}MPI Informatics~~~~~\textsuperscript{2}Saarland Informatics Campus~~~~~\textsuperscript{3}Google~~~~~\textsuperscript{4}University of Pennsylvania
}

\begin{document}

\twocolumn[{
\maketitle
\vspace{-2em}
\centering
Project Page: \href{https://people.mpi-inf.mpg.de/~jianwang/projects/egowholemocap/}{https://people.mpi-inf.mpg.de/~jianwang/projects/egowholemocap/}
\begin{center}
    \captionsetup{type=figure}
    \includegraphics[width=1\textwidth]{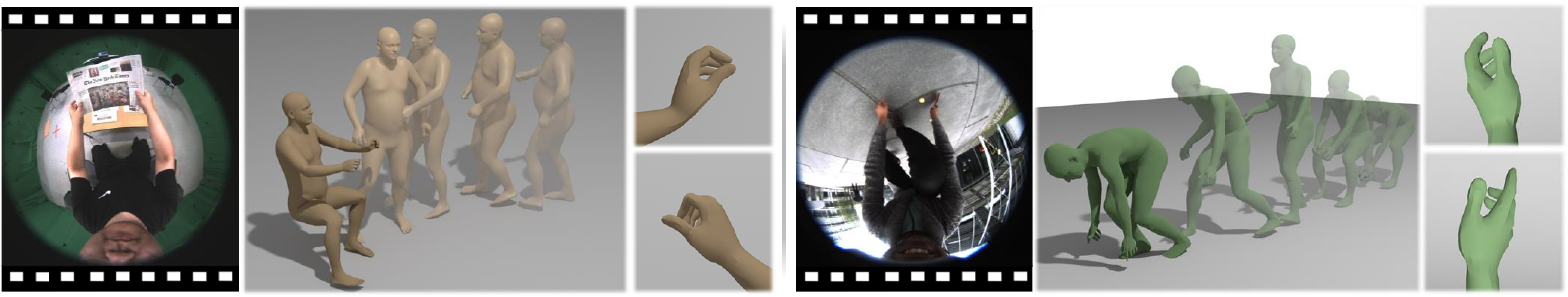}
    \captionof{figure}{From an image sequence captured by a single head-mounted fisheye camera, our method can predict accurate and temporally coherent whole-body motion, including  human body and hand poses. The SMPL-X parameters are obtained using inverse kinematics.
    }
    \label{fig:teaser}
\end{center}
}]

\begin{abstract}
In this work, we explore egocentric whole-body motion capture using a single fisheye camera, which simultaneously estimates human body and hand motion. This task presents significant challenges due to three factors: the lack of high-quality datasets, fisheye camera distortion, and human body self-occlusion. To address these challenges, we propose a novel approach that leverages FisheyeViT to extract fisheye image features, which are subsequently converted into pixel-aligned 3D heatmap representations for 3D human body pose prediction. For hand tracking, we incorporate dedicated hand detection and hand pose estimation networks for regressing 3D hand poses. Finally, we develop a diffusion-based whole-body motion prior model to refine the estimated whole-body motion while accounting for joint uncertainties. To train these networks, we collect a large synthetic dataset, EgoWholeBody, comprising 840,000 high-quality egocentric images captured across a diverse range of whole-body motion sequences. Quantitative and qualitative evaluations demonstrate the effectiveness of our method in producing high-quality whole-body motion estimates from a single egocentric camera. 
\vspace{-1em}
\end{abstract}    
\section{Introduction}
Egocentric 3D human motion estimation using head-mounted devices~\cite{DBLP:journals/tvcg/XuCZRFST19, DBLP:conf/iccv/TomePAB19} has garnered significant traction in recent years, driven by its diverse applications in VR/AR. Immersed in a virtual world, we can traverse virtual environments, interact with virtual objects, and even simulate real-world interactions. To fully capture the intricacies of human motion during such interaction, understanding both body and hand movements is essential. While existing egocentric motion capture methods~\cite{DBLP:journals/tvcg/XuCZRFST19, DBLP:conf/iccv/TomePAB19, wang2021estimating, Wang_2022_CVPR, liu2023egohmr, wang2023scene} focus solely on body motion, neglecting the hands, this work proposes the task of egocentric \emph{whole-body} motion capture, \ie simultaneous estimation of the body motion and hand motion from a single head-mounted fisheye camera (shown in Fig.~\ref{fig:teaser}). This task is extremely challenging due to three factors: 
First, the fisheye image introduces significant distortion, making it difficult for existing networks, which are designed for non-distorted images, to extract features. Second, the egocentric camera perspective frequently leads to the occlusion of body parts, such as the feet and hands, further complicating the task of whole-body motion capture. Lastly, large-scale training data with ground truth annotations for both body and hand poses is absent in existing datasets~\cite{akada2022unrealego,DBLP:journals/tvcg/XuCZRFST19,DBLP:conf/iccv/TomePAB19,Wang_2022_CVPR,liu2023egofish3d}.

In this work, we propose a novel egocentric whole-body motion capture method to address the aforementioned challenges. To effectively address fisheye distortion, we propose \emph{FisheyeViT} for extracting image features, along with a joint regressor employing \emph{pixel-aligned 3D heatmap} for predicting 3D body poses. Instead of attempting to undistort the entire fisheye image, which is impractical due to the fisheye lens's large field of view (FOV), we opt to partition the image into smaller patches aligned with a specific FOV range. This approach enables individual patch-level undistortion and seamlessly aligns with the vision transformer architecture that is employed for extracting the complete image feature map.
We further propose an egocentric 3D pose regressor utilizing 3D heatmap representations. 
Unlike the existing approach~\cite{wang2023scene} that projects image features into 3D space through fisheye reprojection functions and regresses 3D heatmaps with V2V networks~\cite{moon2018v2v}--leading to intricate network learning and high computational complexity--our proposed egocentric pose regressor adopts a simpler approach. It employs deconvolutional layers to obtain pixel-aligned 3D heatmaps. Notably, the voxels in the 3D heatmap directly correspond to pixels in 2D features, subsequently linking to image patches in FisheyeViT. This streamlined approach significantly simplifies network training. 
Joint locations from the pixel-aligned 3D heatmap are finally transformed with the fisheye camera model to obtain the 3D human body poses.
Due to the large size difference between body and hands, we train a hand detection network and a hand pose estimation network to accurately regress 3D hand poses.

To overcome the challenges posed by self-occlusion and improve the accuracy of pose estimation, we propose a novel method for refining the whole-body motion predictions by incorporating temporal context and a motion prior. Our method learns a whole-body motion prior with the diffusion model~\cite{ho2020denoising} from a collection of diverse human motion sequences, capturing intrinsic correlations between hand and body movements. 
Following this, we extract the joint uncertainties from the pixel-aligned 3D heatmap and utilize them to guide the refinement of the whole-body motion.
The joint uncertainties act as indicators of the trustworthiness of the pose regressor's predictions. By conditioning on joints with low uncertainty, our whole-body motion diffusion model selectively refines joints with high uncertainty. This strategy substantially improves the quality of whole-body pose estimations and effectively mitigates the effects of self-occlusion.

In response to the absence of the egocentric whole-body motion capture datasets, we present \emph{EgoWholeBody}, a new large-scale high-quality synthetic dataset. This dataset encompasses a wide range of whole-body motions, comprising over 870k frames,  which significantly surpasses the size of previous egocentric training datasets. EgoWholeBody could serve as a valuable resource for advancing research in egocentric whole-body motion capture.

A thorough evaluation across a range of datasets, including SceneEgo~\cite{wang2023scene}, GlobalEgoMocap~\cite{wang2021estimating} and Mo$^2$Cap$^2$~\cite{DBLP:journals/tvcg/XuCZRFST19}, has demonstrated the remarkable improvements of our method in estimating egocentric whole-body motion compared to previous approaches. This substantiates the effectiveness of our approach in addressing the special challenges encountered in egocentric views, including the fisheye distortion and self-occlusion. 

\noindent In summary, our key contributions are the following:

\begin{itemize}
    \item The first egocentric whole-body motion capture method that predicts accurate and temporarily coherent egocentric body and hand motion;
    \item FisheyeViT for alleviating fisheye camera distortion and pose regressor using pixel-aligned 3D heatmaps for accurate egocentric body pose estimation from a single image;
    \item Uncertainty-aware refinement method based on motion diffusion models for correcting initial pose estimations and predicting plausible motions even under occlusion;
    \item \emph{EgoWholeBody}, a new high-quality synthetic dataset for egocentric whole-body motion capture.
\end{itemize}

\section{Related Works} \label{sec:relatedworks}

\begin{figure*}
\centering
\includegraphics[width=1\linewidth]{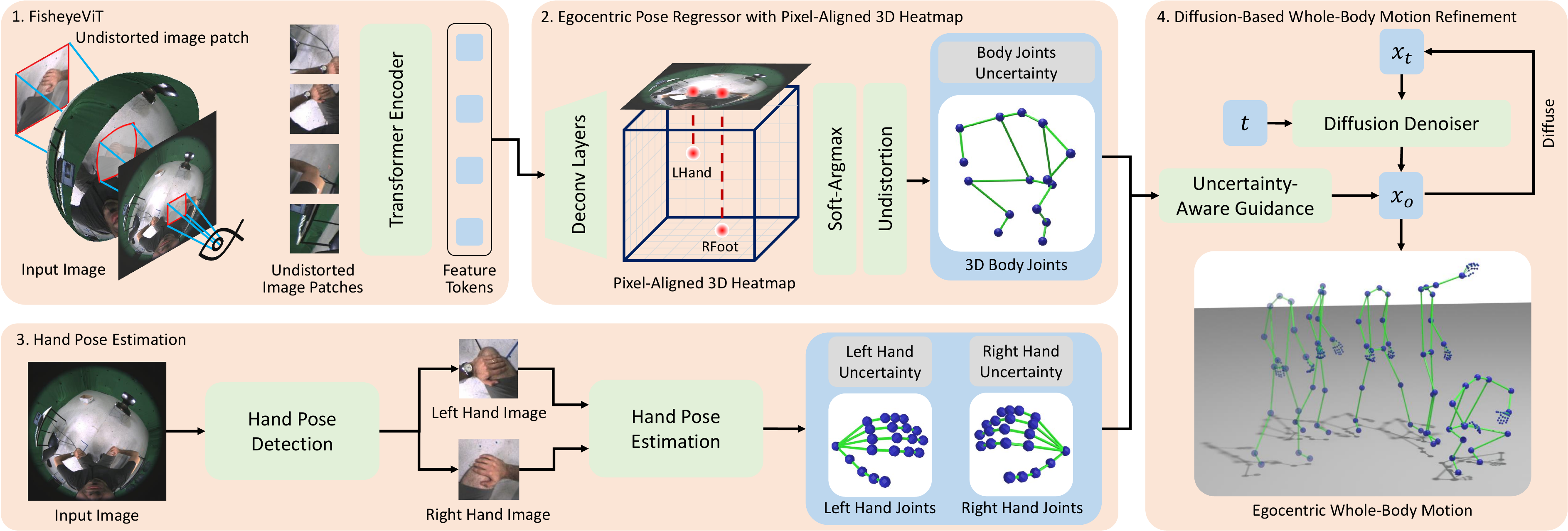}
\caption{Overview of our whole-body motion capture pipeline. We first use FisheyeViT to undistort the input image and generate image feature tokens (\ref{method:fisheyevit}). Next, we use a 1D convolutional network to convert the image features to a pixel-aligned 3D heatmap and use soft-argmax and fisheye camera undistortion function to obtain the 3D body joins positions and uncertainty (\ref{method:pixel_aligned_heatmap}). We further detect the hand location and regress the 3D hand poses from the input image (\ref{method:wholebodypose}). Finally, the estimated hand motion and human body motion are combined and the uncertainty-aware diffusion model is applied to refine the estimated whole-body motion (\ref{method:diffusion}).
\vspace{-1em}
}
\label{fig:framework}
\end{figure*}
\noindent\textbf{Egocentric 3D Human Body Pose Estimation.}
Recently, there has been growing interest in estimating egocentric 3D human poses from body-worn cameras. Some methods ~\cite{li2023ego,jiang2017seeing,ng2020you2me,yuan20183d,yuan2019ego,luo2021dynamics} use front-facing cameras and infer the human body motion from the camera view. However, since the user’s body is often unobserved by the front-facing camera, these methods fail when the human body is not roaming around. Some other methods~\cite{rhodin2016egocap,cha2018towards,akada2022unrealego,zhao2021egoglass,kang2023ego3dpose} use head-mounted down-facing stereo cameras to estimate body poses.
However, stereo camera setups introduce extra burdens of weight and energy consumption. 

Xu~\etal~\cite{DBLP:journals/tvcg/XuCZRFST19} and Tome~\etal~\cite{DBLP:conf/iccv/TomePAB19} introduce the single head-mounted down-facing fisheye camera setup for the egocentric 3D human pose estimation task. 
%
%
Zhang~\etal~\cite{zhang2021automatic} regressed fisheye camera parameters and 3D human pose simultaneously. 
To address the self-occlusion issue, Park~\etal~\cite{park2023domain} leveraged the temporal information with the spatio-temporal self-attention network, and Liu~\etal~\cite{liu2023egohmr} introduced diffusion model to generate 3D human pose conditioned on egocentric image features.
Wang~\etal~\cite{wang2021estimating} and Liu~\etal~\cite{liu2022ego+} combined the SLAM and egocentric pose estimation methods to estimate human body poses in the world coordinate. 
%
Wang~\etal~\cite{Wang_2022_CVPR} and Liu~\etal~\cite{liu2023egofish3d} leverage the synchronized egocentric camera and external cameras to collect large-scale egocentric pose estimation datasets with pseudo-ground truth.
Considering the human-scene interaction, Wang~\etal~\cite{wang2023scene} estimated the scene geometry from the egocentric camera and constrained the 3D human pose with it.

These methods only focus on estimating human body poses while omitting the hand motion, and they still suffer from fisheye camera distortion since they directly put the highly distorted fisheye images into the neural network. Our proposed method can capture whole-body motion and resolve the fisheye camera distortion issue with the FisheyeViT and pixel-aligned 3D heatmap.

\noindent\textbf{Whole-Body 3D Pose Estimation.}
Whole-body 3D pose estimation aims to estimate the 3D human body, face, and hands parameters from input images. This task is crucial for many applications, e.g., modeling human activities and human-scene interactions.
Some methods~\cite{pavlakos2019expressive,xiang2019monocular} fit the 2D body joints estimated from images with optimization algorithms, while these methods suffer from high computation overhead and can fall into local optima.
Some other learning-based methods~\cite{sun2022learning,zhou2021monocular,feng2021collaborative,choutas2020monocular,rong2021frankmocap,lin2023one,cai2023smpler} use the neural network to regress the SMPL-X~\cite{pavlakos2019expressive} parameters from input images. For example, ExPose~\cite{choutas2020monocular} introduced body-driven attention to extract face and hand crops and used a refinement module to regress whole-body pose. 
OSX~\cite{lin2023one} proposed a one-stage pipeline for whole-body mesh recovery without separate networks for each part. 
SMPLer-X~\cite{cai2023smpler} propose a foundation model for whole-body pose estimation trained with the large model and big data.

Though much progress has been made on whole-body pose estimation from an external view, the task from an egocentric view is still unexplored. In this paper, we introduce the first whole-body 3D pose estimation method from a single egocentric image and also incorporate temporal information with diffusion-based motion refinement.

\noindent\textbf{Diffusion Models for Pose Estimation.}
Recently, some methods ~\cite{holmquist2023diffpose, gong2023diffpose, choi2022diffupose, shan2023diffusion, foo2023distribution, ci2023gfpose} have effectively applied Denoising Diffusion Probabilistic Models (DDPM)~\cite{ho2020denoising} to human pose estimation tasks.
Building on the success of motion diffusion models in human pose estimation, many methods have extended this approach to egocentric pose estimation, where the human body is only partially visible from RGB cameras or VR sensors. Zhang~\etal's work~\cite{zhang2023probabilistic} uses a diffusion model to generate realistic human poses considering scene geometry.  AGROL~\cite{du2023avatars} generates body motion based on head and hand 6D pose estimates from a VR headset.  EgoEgo~\cite{li2023ego} estimates head poses from a head-mounted front-facing camera and uses them to generate body poses. EgoHMR~\cite{liu2023egohmr} extracts image features and uses them as a condition for the diffusion denoising process.

However, the aforementioned pose estimation methods train the \emph{conditioned} diffusion model with image features or IMU signals. This cannot be generalized since the trained network only accepts one specific condition format and is inclined to learn domain-specific distributions of condition features. ZeDO~\cite{jiang2023back} tackles this issue with a zero-shot diffusion-based optimization approach that doesn't require training with 2D-3D or image-3D pairs. Our method leverages the uncertainty value given by the single-frame pose estimation network and refines the initial motion estimation with the uncertainty of each joint. Moreover, different from previous methods that only focus on human body motion, we train a whole-body motion diffusion model to construct the relationship between hand and body motion.

\section{Method} \label{method}
In this section, we propose a new method for predicting accurate egocentric whole-body poses from egocentric image sequences. An overview of our approach is shown in Fig.~\ref{fig:framework}. 

\subsection{Single Image Based Egocentric Pose Estimation} \label{method:single_image_pose}

\subsubsection{FisheyeViT} \label{method:fisheyevit}

\begin{figure}
\centering
\includegraphics[width=0.96\linewidth]{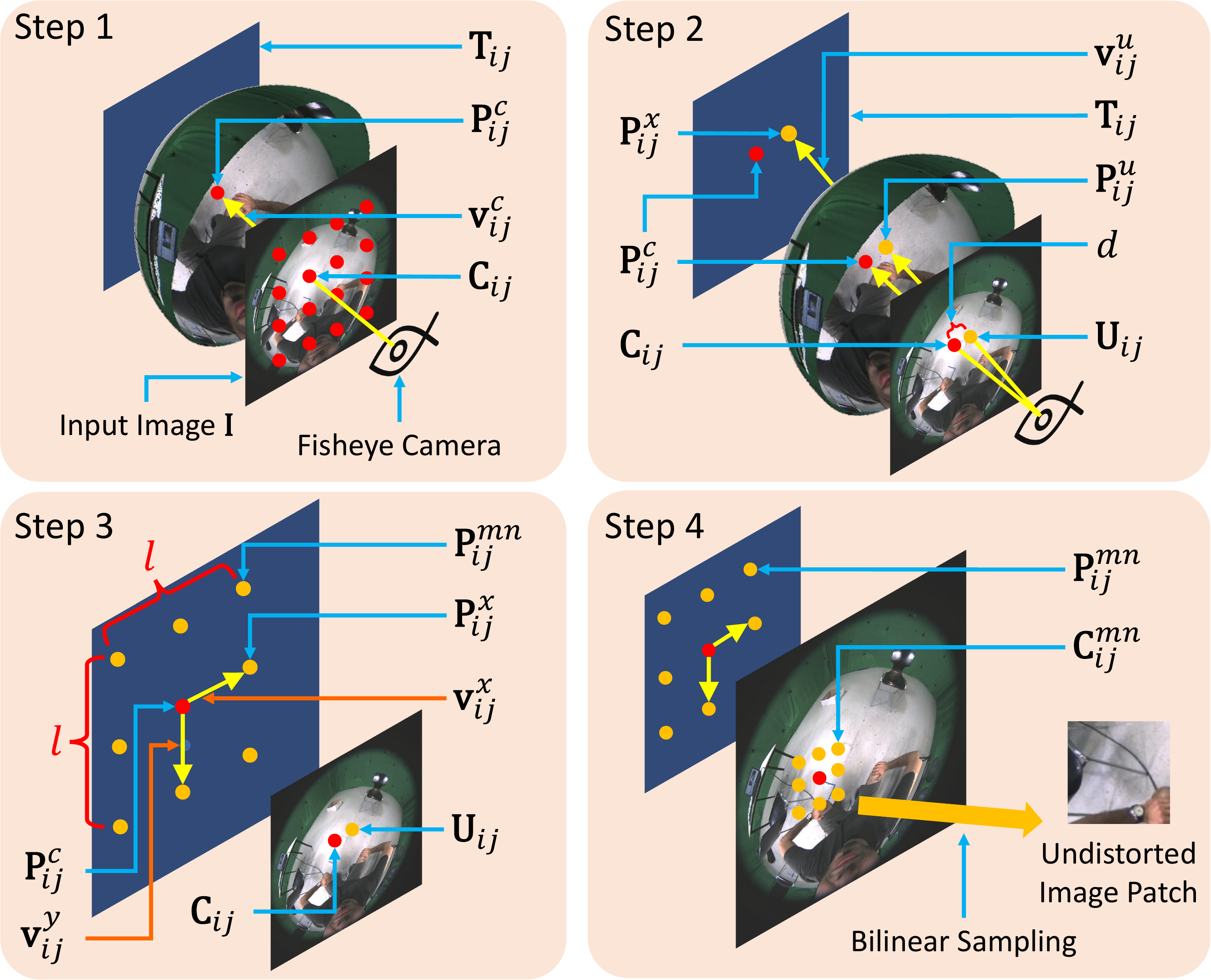}
\caption{The detailed illustration of FisheyeViT (Sec.~\ref{method:fisheyevit}).
\vspace{-1em}
}
\label{fig:fisheye_vit_inline}
\end{figure}

In this section, we introduce FisheyeViT, which is specially designed to alleviate the fisheye distortion issue. Instead of undistorting the entire fisheye image, we extract undistorted image patches from the fisheye image and then fit these patches as tokens into the transformer network~\cite{dosovitskiy2020image}. 
To get the undistorted patches, we first warp the fisheye image to a unit semi-sphere, then get the patches with the gnomonic projection (see \cref{fig:framework}).
The FisheyeViT can be split into five steps, the first four of which are illustrated in Fig.~\ref{fig:fisheye_vit_inline}.

\textbf{Step 1.} Given an input image $\mathbf{I}$ with size $H\times W$, we first evenly sample $N\times N$ patch center points: $\{\mathbf{C}_{ij} = (u_i, v_j) = \left(\frac{H}{N}(i + \frac{1}{2}), \frac{W}{N}(j + \frac{1}{2})\right) | i,j \in 0, ..., N-1\}$. Then, the patch center points $\mathbf{C}_{ij}$ are projected onto a unit sphere with the fisheye reprojection function: 
$\mathbf{P}^c_{ij} = (x^c_{ij}, y^c_{ij}, z^c_{ij}) = \mathcal{P}^{-1}(u_i, v_j, 1)$. The fisheye camera model is described in Sec.~\ref{supp:fisheye_model} of the supplementary material. Given a point $\mathbf{P}^c_{ij}$ on the unit sphere, the tangent plane $\mathbf{T}_{ij}$ that passes through the point is defined by the normal vector $\mathbf{v}^c_{ij} = (x^c_{ij}, y^c_{ij}, z^c_{ij})$. In the following steps, we implement the gnomonic projection by sampling grid points in the plane and projecting them back onto the fisheye image.

\textbf{Step 2.} In this step, we determine the orientation of the grid points in the tangent plane, ensuring that the grid points from different tangent planes $\mathbf{T}_{ij}$ have the same orientation when projected back onto the fisheye image. To achieve this, we select a 2D point $\mathbf{U}_{ij} = (u_i + d, v_j)$ in the fisheye image space that is $d$ pixels to the right of the patch center point and project it to the unit sphere using the fisheye reprojection function: $\mathbf{P}^u_{ij} = (x^u_{ij}, y^u_{ij}, z^u_{ij}) = \mathcal{P}^{-1}(u_i + d, v_j, 1)$. We then calculate the intersection point $\mathbf{P}^x_{ij}$ between the vector $\mathbf{v}^u_{ij} = (x^u_{ij}, y^u_{ij}, z^u_{ij})$ that is passing the origin and the tangent plane $\mathbf{T}_{ij}$: 
\begin{equation}
    \mathbf{P}^x_{ij} = \frac{\left<\mathbf{P}^c_{ij}, \mathbf{v}^c_{ij} \right>}{\left<\mathbf{v}^u_{ij}, \mathbf{v}^c_{ij} \right> } \mathbf{v}^u_{ij} = \frac{1}{\left<\mathbf{v}^u_{ij}, \mathbf{v}^c_{ij} \right> } \mathbf{v}^u_{ij},
\end{equation}
where $\left<\cdot,\cdot\right>$ denotes the inner product.

\textbf{Step 3.} Based on the center point $\mathbf{P}^c_{ij}$ and intersection point $\mathbf{P}^x_{ij}$ on the tangent plane $\mathbf{T}_{ij}$, we build a coordinate system with the $x$ axis: $\mathbf{v}^x_{ij} = \text{Norm}(\mathbf{P}^x_{ij} - \mathbf{P}^c_{ij})$, the $z$ axis: $\mathbf{v}^z_{ij} = \text{Norm}(\mathbf{v}^c_{ij})$ and the $y$ axis: $\mathbf{v}^y_{ij} = \mathbf{v}^z_{ij} \times \mathbf{v}^x_{ij}$, where $\text{Norm}$ denotes the normalize operation. We grid-sample $M \times M$ points in a $l \times l$ square on the $x\text{-}y$ plane: 
\begin{equation}
    \{\mathbf{P}^{mn}_{ij} = \mathbf{P}^c_{ij} + (l\frac{m}{M} \mathbf{v}^x_{ij}, l\frac{n}{M} \mathbf{v}^y_{ij})\}
\end{equation}
where $m, n \in -\frac{1}{2}(M-1), ..., -\frac{3}{2}, -\frac{1}{2}, \frac{1}{2}, \frac{3}{2}, ... \frac{1}{2}(M-1)$. 

\textbf{Step 4.} The points $\mathbf{P}^{mn}_{ij}$ are projected back to the fisheye image with the fisheye projection function: $\mathbf{C}^{mn}_{ij} = \mathcal{P}(\mathbf{P}^{mn}_{ij})$. We then apply bilinear sampling to obtain the colors at points $\mathbf{C}^{mn}_{ij}$ of the input image $\mathbf{I}$, yielding the undistorted image patch $\mathbf{I}^{\text{undis}}_{ij}$. 
Please also see the supplementary video for a visual demonstration of undistorted image patches and their movement on the fisheye image.

\textbf{Step 5.} The image patches $\{\mathbf{I}^{\text{undis}}_{ij}\}$ are sent to a ViT transformer network~\cite{dosovitskiy2020image} to obtain the feature tokens $\{\mathbf{F}_{ij}\}$. The feature token is further reshaped in $i \times j$ matrix and obtain the image feature $\mathbf{F}$. In the FisheyeViT, we empirically chose $N=16; M=16; d=8; l=0.2m$ given the image size $H=W=256$.

Note that $\mathbf{C}^{mn}_{ij}$ is independent of the image $\mathbf{I}$. This means that, given a fixed fisheye camera model, we can precompute $\mathbf{C}^{mn}_{ij}$ for all combinations of $m, n$ and $i, j$ in advance. This significantly speeds up both the training and evaluation processes. Furthermore, the number and dimensions of image patches $\{\mathbf{I}^{\text{undis}}_{ij}\}$ match exactly with those in the traditional ViT network. This compatibility allows us to finetune existing ViT networks on our egocentric datasets.
Our sampling strategy ensures that each image patch $\mathbf{I}^{\text{undis}}_{ij}$ corresponds to the same FOV range in the fisheye camera. In our ablation study in Sec.~\ref{ablation_study}, we show that FisheyeViT enhances the performance of the pose estimation network when applied to egocentric fisheye images.

\subsubsection{Pose Regressor with Pixel-Aligned 3D Heatmap} \label{method:pixel_aligned_heatmap}

After collecting image features with FisheyeViT, we utilize a 3D heatmap-based network to estimate the body poses. The existing 3D heatmap-based pose regressors~\cite{sun2018integral, Moon_2022_CVPRW_Hand4Whole} are designed for the weak-perspective cameras and predict the 3D heatmap in $xyz$ space.
Directly applying these regressors will result in misalignment between 3D heatmap features in $xyz$ space and 2D image features in the fisheye image space. Therefore, we introduce a novel egocentric pose regressor that relies on the pixel-aligned 3D heatmap, tailored to address the needs of fisheye cameras. The idea is to regress the 3D heatmap in $uvd$ space rather than traditional  $xyz$ space, where $uv$ corresponds to the fisheye image $uv$ space.
Specifically, given a feature map $\mathbf{F} \in \mathbb{R}^{C \times N \times N}$, where $C$ is the channel number, $N$ is feature map height and width, we firstly use two deconvolutional layers to convert the feature map $\mathbf{F}$ into shape $(D_h \times J, H_h, W_h)$, and further reshape it to pixel-aligned 3D heatmap $\mathbf{H}\in \mathbb{R}^{J \times D_h \times H_h \times W_h}$, where $J$ is the joint number and $D_h$, $H_h$, $W_h$ is the 3D heatmap depth, height and width. The illustration of pixel-aligned 3D heatmap is shown in Fig.~\ref{fig:framework}. Next, we obtain the max-value positions $\mathbf{\tilde{J}}_b=\{(u_i, v_i, d_i) \mid i\in 0, 1, 2,..., J\}$ from $\mathbf{H}$ by the differentiable soft-argmax operation~\cite{sun2018integral}. Here, we note that $u_i$ and $v_i$ correspond to the uv-coordinate of the 3D body joint projected in the fisheye image space, and $d_i$ denotes the distance of the joint to the fisheye camera. Finally, the 3D body joints $\mathbf{\hat{J}}_b=\{(x_i, y_i, z_i) \mid i\in 0, 1, 2,..., J\}$ are recovered with the fisheye reprojection function: $(x_i, y_i, z_i) = \mathcal{P}^{-1}(u_i, v_i, d_i)$. The predicted body pose $\mathbf{\hat{J}}_b$ is finally compared with the ground truth body pose $\mathbf{J}_{b}$ with the MSE loss. 
By first regressing 3D body poses in $uvd$ space and then reprojecting it, we ensure that the 3D heatmap is pixel-aligned with the end-to-end training.

With the pixel-aligned heatmap, our proposed 3D pose regressor solves problems in all three types of previous egocentric joint regressors. 
First, Mo$^2$Cap$^2$~\cite{DBLP:journals/tvcg/XuCZRFST19} employs separate networks to predict 2D joint positions and joint distances. However, this method can yield unrealistic joint estimations because small errors in 2D joints can result in large errors in 3D joints due to the projection effect. 
Second, $x$R-egopose~\cite{DBLP:conf/iccv/TomePAB19} and EgoHMR~\cite{liu2023egohmr} directly regress the 3D joint positions. However, this method is agnostic to the fisheye camera parameters, making it suitable only for a specific camera configuration (\eg, camera parameters, head-mounted position, \etc). 
Third, SceneEgo~\cite{wang2023scene} projects 2D features into 3D voxel space and uses a V2V network to regress 3D poses. Because of these, the SceneEgo method suffers from low accuracy and large computation overhead.
Different from previous methods, our pose regressor with pixel-aligned 3D heatmap is versatile and efficient since it directly estimates 3D joints while also incorporating an explicitly parametrized fisheye camera model. 
Moreover, it can preserve the uncertainty of the estimated joints, which will be used in our uncertainty-aware motion refinement method (Sec.~\ref{method:uncertainty_aware}). Detailed comparison with other pose prediction heads is shown in Table~\ref{table:ablation_study}.

\subsubsection{Egocentric Hand Pose Estimation} \label{method:wholebodypose}

In this section, we first train a network to detect hand pose locations and then train a 3D hand pose estimation network to regress 3D hand poses. Then, we describe how to integrate the estimated hand and body poses.

\noindent\textbf{Hand Detection.} \label{method:hand_detection}
Given an input image $\mathbf{I}$, we finetune the HRNet~\cite{wang2020deep} network to regress the 2D hand poses of left hand $\mathbf{J}^{2d}_{lh}$ and right hand $\mathbf{J}^{2d}_{rh}$. From the hand poses, we obtain the center point of left hand $\mathbf{C}_{lh}$ and right hand $\mathbf{C}_{rh}$, along with the bounding box sizes, $d_{lh}$ and $d_{rh}$. We use our approach described in Sec.~\ref{method:fisheyevit} to compute undistorted image patches of left $\mathbf{I}_{lh}$ and right hands $\mathbf{I}_{rh}$.

\noindent\textbf{Hand Pose Estimation.} \label{method:hand_pose}
Given the cropped image $\mathbf{I}_{lh}$ or $\mathbf{I}_{rh}$, we regress the 3D hand poses $\mathbf{\hat{J}}^{loc}_{lh}$ and $\mathbf{\hat{J}}^{loc}_{rh}$ with the Hand4Whole~\cite{Moon_2022_CVPRW_Hand4Whole} network, which is fine-tuned on our EgoFullBody dataset.

\noindent\textbf{Integration of Body and Hand Poses.}
It is not straightforward to integrate the hand poses with the body pose in the egocentric camera view primarily due to the fisheye camera's perspective effects. Take the left hand as an example. Following Step 3 in Sec.~\ref{method:fisheyevit}, we establish a local coordinate system on the tangent plane of the left-hand image with XYZ axes as follows: $x: \mathbf{v}^x_{lh}$; $y: \mathbf{v}^y_{lh}$; $z: \mathbf{v}^z_{lh}$. We define a rotation matrix, denoted as $\mathbf{R}$, that represents the transformation between the root coordinate system and the local coordinate system on the tangent plane. The estimated hand pose is first rotated with the rotation matrix $\mathbf{\hat{J}}_{lh} = \mathbf{R}\mathbf{\hat{J}}^{loc}_{lh}$ and then translated to align the wrist location of the human body. This same process is also applied to the right hand to get the right hand pose $\mathbf{\hat{J}}_{rh}$. The whole-body joints $\mathbf{\hat{J}}$ are obtained by combining $\mathbf{\hat{J}}_{b}$,  $\mathbf{\hat{J}}_{lh}$, and $\mathbf{\hat{J}}_{rh}$. The uncertainty of whole-body joints $\mathbf{\hat{U}}$ is also obtained from the maximal value of the 3D heatmap in pose estimation modules.

\subsection{Diffusion-Based Motion Refinement} \label{method:diffusion}

We notice that the single-frame estimations in Sec.~\ref{method:single_image_pose} suffer from inaccuracies and temporal instabilities. In this section, we propose a diffusion-based motion refinement method to tackle this problem. We first learn the whole-body motion prior with the motion diffusion model in Sec.~\ref{method:wholebody_motion_diffusion} and then introduce an uncertainty-aware zero-shot motion refinement method in Sec.~\ref{method:uncertainty_aware}.

\subsubsection{Whole-Body Motion Diffusion Model} \label{method:wholebody_motion_diffusion}


We follow DDPM~\cite{ho2020denoising} as our diffusion approach to capture the whole-body motion prior $q(\mathbf{x})$. DDPM learns a distribution of whole-body motion $\mathbf{x}$ through a forward diffusion process and an inverse denoising process. 
The forward diffusion process is a Markov process of adding Gaussian noise over $t \in \{0, 1, ..., T-1\}$ steps:
\begin{equation}
    q(\mathbf{x}_t|\mathbf{x}_{t-1}) = \mathcal{N}(\sqrt{\alpha_t}\mathbf{x}_{t-1}, (1 - \alpha_t)I)
\end{equation}
where $\mathbf{x}_t$ denotes the whole-body motion sequence at step $t$, the variance $(1-\alpha_t) \in (0, 1]$ denotes a constant hyperparameter increases with $t$. 

The inverse process uses a denoising network $D(\cdot)$ to remove the added Gaussian noise at each time step $t$. Here we use the transformer-based framework in EDGE~\cite{tseng2023edge} as the motion-denoising network $D(\cdot)$. We follow Ramesh~\etal's work~\cite{ramesh2022hierarchical} to make the network predict the original signal itself, \ie $\mathbf{\hat{x}}_0 = D(\mathbf{x}_t, t)$ and train it with the simple objective~\cite{ho2020denoising}:
\begin{equation} \label{eq:diffusion_simple_objective}
    \mathcal{L}_{\text{simple}} = E_{\mathbf{x}_0 \sim q(\mathbf{x}_0), t \sim [1, T]}\left[ \vert\vert\mathbf{x}_0 - D(\mathbf{x}_t, t)\vert\vert^2_2 \right]
\end{equation}

\subsubsection{Uncertainty-Aware Motion Refinement} \label{method:uncertainty_aware}

Given the learned whole-body motion prior, we leverage the uncertainty value for each pose to guide the diffusion denoising process with the classifier-guided diffusion sampling~\cite{dhariwal2021diffusion}. Given an initial sequence of whole-body pose estimation $\mathbf{x}_e = \{\mathbf{\hat{J}}_i\}$ and the uncertainty value for each pose $\mathbf{u} = \{\mathbf{\hat{U}}_i\}$, where $i$ denotes the $i$th pose in the sequence, we keep the joints with low uncertainty but use the diffusion model to generate joints with high uncertainty conditioned on the low-uncertainty joints. Specifically, in the $t$th sampling step of the diffusion process, the denoising network predicts $\mathbf{\hat{x}}_0 = D(\mathbf{x}_t, t)$, which is noised back to $\mathbf{x}_{t-1}$ by sampling from the Gaussian distribution:
\begin{equation}\label{eq:diffusion_refinement}
    \mathbf{x}_{t-1} \sim \mathcal{N}(\mathbf{\hat{x}}_0 + \mathbf{w}(\mathbf{x}_e - \mathbf{\hat{x}}_0), \Sigma_t)
\end{equation}
where $\Sigma_t$ is a scheduled Gaussian distribution in DDPM~\cite{ho2020denoising} and $\mathbf{w}$ controls the weight of a specific joint between the predicted motion $\mathbf{\hat{x}}_0$ and the estimated motion $\mathbf{x}_e$. Generally, we expect $\mathbf{w} \rightarrow \overrightarrow{0}$ when $t \rightarrow 0$ such that the temporal stability is guaranteed through the generation of the denoising process, and $\mathbf{w} \rightarrow \overrightarrow{1}$ when $t \rightarrow T$ such that the denoising process is initialized by the estimated motion $\mathbf{x}_e$. We also expect that $w_{ij} = \mathbf{w}[i][j]$, which is the weight of $j$th joints in the $i$th pose, is smaller when the uncertainty value $u_{ij} = \mathbf{u}[i][j]$ of the $j$th joints in the $i$th pose is large. Based on this requirement, we design $\mathbf{w}$ as:
\begin{equation} \label{eq:uncertainty_weight}
    \mathbf{w} = 1 / \left(1 + e^{-k(t-T\mathbf{u})}\right)
\end{equation}
where $T$ is the overall diffusion steps, $k$ is a hyperparameter which is empirically set to $0.1$.
From the experimental results in Sec.~\ref{experiments}, we demonstrate the effectiveness of uncertain-aware motion refinement and our uncertainty-guided diffusion sampling strategy.

\section{EgoWholeBody Dataset} \label{method:synthetic_dataset}

In this section, we introduce EgoWholeBody, a large-scale high-quality synthetic dataset built for the task of egocentric whole-body motion capture. The EgoWholeBody dataset is organized into two sections. The first part, containing over 700k frames, is rendered with 14 different rigged Renderpeople~\cite{renderpeople} models driven by 2367 Mixamo~\cite{mixamo} motion sequences. Note that the Mixamo~\cite{mixamo} motions used in this part mainly focus on body movements, lacking diversity in hand motions. The second part of our dataset focuses on more complex hand motions and contains 170k frames with the SMPL-X model. This data is constructed from 24 different shapes and textures, driven by 200 motion sequences selected from the GRAB dataset~\cite{taheri2020grab} and 62 motion sequences from the TCDHandMocap dataset~\cite{hoyet2012sleight}. 

During the rendering process, we first attach a virtual fisheye camera to the forehead of human body models and render the images, semantic labels, and depth map with Blender~\cite{blender}. 
Our dataset is larger and more diverse than previous egocentric training datasets--see Sec.~\ref{supp:dataset_comparison} in the supplementary material for a detailed comparison. 
\section{Experiments} \label{experiments}

\begin{figure*}
\centering
\includegraphics[width=0.99\linewidth]{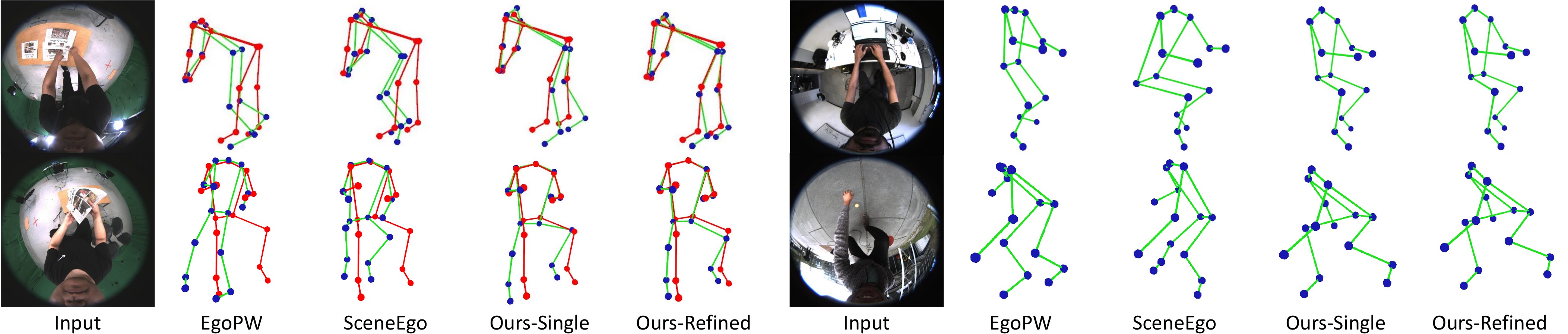}
\caption{Qualitative comparison on human body pose estimations between our methods and the state-of-the-art egocentric pose estimation methods on in-the-studio (left column) and in-the-wild scenes (right column). The red skeleton is the ground truth while the green skeleton is the predicted pose. Our methods predict more accurate body poses compared with EgoPW~\cite{Wang_2022_CVPR} and SceneEgo~\cite{wang2023scene}.
\vspace{-1em}
}
\label{fig:results}
\end{figure*}

\begin{figure*}
\centering
\includegraphics[width=0.99\linewidth]{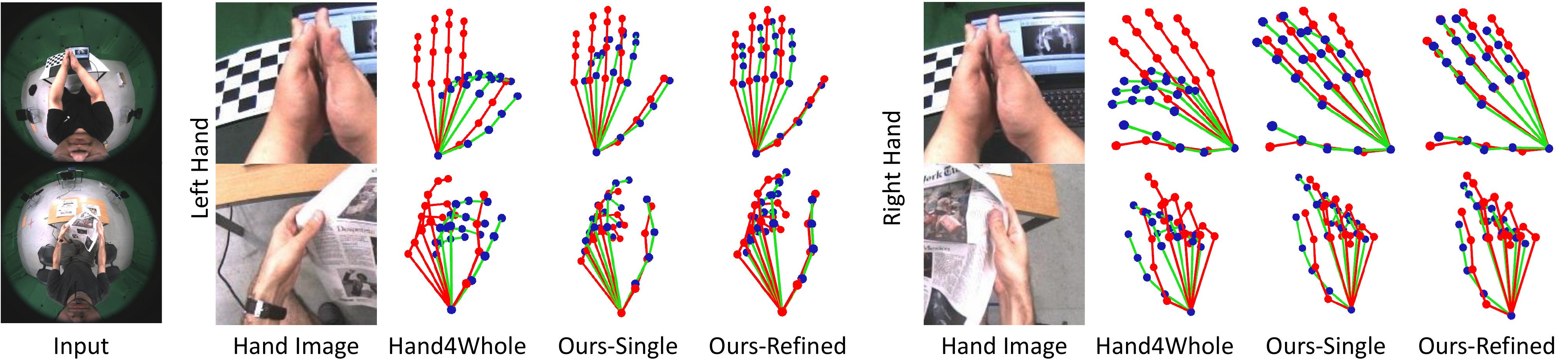}
\caption{Qualitative comparison on human hand pose estimations between our methods and the state-of-the-art third-view pose estimation methods. Our single-view and refined hand poses are more accurate than the poses from Hand4Whole~\cite{Moon_2022_CVPRW_Hand4Whole} method. The red skeleton is the ground truth while the green skeleton is the predicted pose. 
\vspace{-1em}
}
\label{fig:results_hand}
\end{figure*}

\subsection{Datasets and Evaluation Metrics}

\noindent\textbf{Training Datasets.} \label{dataset:training}
To train our body pose estimation module (Sec.~\ref{method:fisheyevit} and Sec.~\ref{method:pixel_aligned_heatmap}), we use our EgoWholeBody dataset and the EgoPW dataset~\cite{wang2021estimating}. 
Additionally, the EgoWholeBody dataset is used to train the hand pose estimation module in Sec.~\ref{method:wholebodypose}.
For training the whole-body diffusion model (Sec.~\ref{method:diffusion}), we utilize a combined motion capture dataset that includes EgoBody~\cite{zhang2022egobody}, Mixamo~\cite{mixamo}, TCDHandMocap dataset~\cite{hoyet2012sleight} and GRAB dataset~\cite{taheri2020grab}.

\noindent\textbf{Evaluation Datasets.} 
In our experiment, we evaluate our methods on three datasets: the GlobalEgoMocap test datasets~\cite{wang2021estimating}, the Mo$^2$Cap$^2$ test dataset~\cite{DBLP:journals/tvcg/XuCZRFST19} and the SceneEgo test dataset~\cite{wang2023scene}. The details of these datasets are shown in Sec.~\ref{supp:evaluation_dataset} of supplementary materials. 
Note that evaluating whole-body poses requires accurate annotations for human hands, which is absent in these datasets. To resolve the issue, we request the multi-view videos of the SceneEgo test dataset~\cite{wang2023scene} from the authors and use a multi-view motion capture system to obtain the hand motion. The hand pose annotations will be made public available.

\noindent\textbf{Evaluation Metrics.}
To evaluate the precision of human body poses on the SceneEgo test dataset~\cite{wang2023scene}, we use MPJPE and PA-MPJPE. For the GlobalEgoMocap test dataset~\cite{wang2021estimating} and Mo$^2$Cap$^2$ test dataset~\cite{DBLP:journals/tvcg/XuCZRFST19}, where egocentric camera poses are unavailable, we evaluate PA-MPJPE and BA-MPJPE.
For hand pose accuracy, we align the predicted and ground truth hand poses at the root position, followed by computing MPJPE and PA-MPJPE. Detailed explanations of these metrics are in Sec.~\ref{supp:evaluation_metrics} of the supplementary materials. All reported metrics are in millimeters.

\subsection{Comparisons on Whole-Body Pose Estimation} \label{experiments:main_comparison}
For a fair comparison with existing methods focusing solely on body or hand pose, we split our evaluation into two parts, reporting results of body poses in Table~\ref{table:main} and hand pose in Table~\ref{table:hand}.
We first compare the accuracy of human body poses with state-of-the-art methods, including EgoPW~\cite{Wang_2022_CVPR}, SceneEgo~\cite{wang2023scene}, EgoHMR~\cite{liu2023egohmr} and Ego-STAN~\cite{park2023domain}. Since our motion refinement method incorporates random Gaussian noise, we generate five samples and calculate the average MPJPE values. The standard deviation of the results is low ($<$ 0.01mm) and is discussed in Sec.~\ref{supp:std} of supplementary materials. Results are presented in Table~\ref{table:main}, where our single-frame results are labeled as ``Ours-Single'' and our refinement results are labeled as ``Ours-Refined''. Across all three evaluation datasets, our single-frame body pose estimation method outperforms all previous methods by a large margin. Our diffusion-based motion refinement method can further improve the accuracy of body poses estimated by the single-frame methods.

To evaluate the accuracy of our hand pose estimation method, we first crop the hand images with the hand detection method in Sec.~\ref{method:wholebodypose}. Then we show the results of our single-frame hand pose estimation (labeled as ``Ours-Single'') and whole-body motion refinement methods (labeled as ``Ours-Refined'') in Table~\ref{table:hand}. Our single-frame hand pose estimation method outperforms the state-of-the-art method Hands4Whole~\cite{Moon_2022_CVPRW_Hand4Whole}, demonstrating the effectiveness of training the network on our EgoWholeBody dataset. Our whole-body motion refinement method can also enhance the accuracy of hand motion.


For a qualitative comparison, we compare the body and hand poses of our method with existing methods on the SceneEgo dataset and the in-the-wild EgoPW~\cite{Wang_2022_CVPR} evaluation sequences. The results are shown in Fig.~\ref{fig:results} and Fig.~\ref{fig:results_hand}, showing that our method can predict high-quality whole-body poses from an egocentric camera. Please refer to our supplementary video for more qualitative evaluation results.

We observe that previous methods~\cite{DBLP:journals/tvcg/XuCZRFST19, DBLP:conf/iccv/TomePAB19, wang2021estimating, Wang_2022_CVPR, wang2023scene} use training datasets different from each other. To ensure a fair comparison in body pose estimation methods, we re-train previous methods with our training datasets in Sec.~\ref{dataset:training} and show the results in Sec.~\ref{supp:full_comparison} of the supplementary materials.

\begin{table}
\begin{center}
\small
\begin{tabularx}{0.47\textwidth} { 
   >{\raggedright\arraybackslash}X 
   >{\centering\arraybackslash}X 
   >{\centering\arraybackslash}X  }
\hlineB{2.5}
Method & MPJPE & PA-MPJPE \\
\hline
\multicolumn{2}{l}{\textbf{SceneEgo test dataset~\cite{wang2023scene}}} \\
\hline
EgoPW~\cite{Wang_2022_CVPR} & 189.6 & 105.3 \\
SceneEgo~\cite{wang2023scene} & 118.5 & 92.75 \\
\hline
Ours-Single & \underline{64.19} & \underline{50.06} \\
Ours-Refined$^\dagger$ & \textbf{57.59} & \textbf{46.55} \\
\hlineB{2.5}
Method & PA-MPJPE & BA-MPJPE \\
\hline
\multicolumn{2}{l}{\textbf{GlobalEgoMocap test dataset~\cite{wang2021estimating}}} \\
\hline
EgoPW~\cite{Wang_2022_CVPR} & 81.71 & 64.87 \\
EgoHMR~\cite{liu2023egohmr} & 85.80 & -- \\
SceneEgo~\cite{wang2023scene} & 76.50 & 61.92 \\
\hline
Ours-Single & \underline{68.59} & \underline{55.92} \\
Ours-Refined$^\dagger$ & \textbf{65.83} & \textbf{53.47} \\
\hlineB{2.5}
\multicolumn{2}{l}{\textbf{Mo$^2$Cap$^2$ test dataset~\cite{DBLP:journals/tvcg/XuCZRFST19}}} \\
\hline
EgoPW~\cite{Wang_2022_CVPR} & 83.17 & 64.33 \\
Ego-STAN$^\dagger$~\cite{park2023domain} & 102.4 & -- \\
SceneEgo~\cite{wang2023scene} & 79.65 & 62.82 \\
\hline
Ours-Single & \underline{74.66} & \underline{59.26} \\
Ours-Refined$^\dagger$ & \textbf{72.63} & \textbf{57.12} \\
\hlineB{2.5}
\end{tabularx}
\end{center}
\vspace{-1em}
\caption{
Egocentric human body pose accuracy of our method on three test datasets. Our method outperforms all previous state-of-the-art methods. $^\dagger$ denotes the temporal-based methods. 
}
\vspace{-0.5em}
\label{table:main}
\end{table}

\begin{table}
\begin{center}
\small
\begin{tabularx}{0.47\textwidth} { 
   >{\raggedright\arraybackslash}X 
   >{\centering\arraybackslash}X 
   >{\centering\arraybackslash}X  }
\hlineB{2.5}
Method & MPJPE & PA-MPJPE \\
\hline
Hand4Whole~\cite{Moon_2022_CVPRW_Hand4Whole} & 49.66 & 13.85 \\
Ours-Single & 23.63 & 9.59 \\
Ours-Refined & \textbf{19.37} & \textbf{9.05} \\
\hlineB{2.5}
\end{tabularx}
\end{center}
\vspace{-1em}
\caption{Egocentric hand pose accuracy of our method on SceneEgo test dataset. Our method outperforms the state-of-the-art Hand4Whole method~\cite{Moon_2022_CVPRW_Hand4Whole}. 
\vspace{-0.5em}
}
\label{table:hand}
\end{table}

\subsection{Ablation Study} \label{ablation_study}

\begin{table}
\begin{center}
\small
\begin{tabularx}{0.47\textwidth} { 
   >{\raggedright\arraybackslash}X 
   >{\centering\arraybackslash\hsize=.5\hsize}X 
   >{\centering\arraybackslash\hsize=.5\hsize}X  }
\hlineB{2.5}
Method & MPJPE & PA-MPJPE \\
\hline
\multicolumn{2}{l}{\textbf{Body Pose Results}} \\
\hline
w/o EgoWholeBody  & 75.10 & 58.62 \\
w/o FisheyeViT  & 67.36 & 53.44 \\
w/ Mo$^2$Cap$^2$~\cite{DBLP:journals/tvcg/XuCZRFST19} head  & 87.47 & 65.10 \\
w/ $x$R-egopose~\cite{DBLP:conf/iccv/TomePAB19} head  & 116.5 & 95.78 \\
w/ SceneEgo~\cite{wang2023scene} head  & 77.73 & 62.69 \\
Ours-Single & \textbf{64.19} & \textbf{50.06} \\
\hline
w/ GlobalEgoMocap$^\dagger$ & 69.83 & 56.73 \\
w/o uncert. guidance$^\dagger$ & 62.16 & 48.40 \\
Only body diffusion & 58.95 & 47.03 \\
Ours-Refined$^\dagger$ & \textbf{57.59}  & \textbf{46.55} \\
\hline
\multicolumn{2}{l}{\textbf{Hand Pose Results}} \\
\hline
Only hand diffusion & 21.69 & 9.24 \\
Ours-Refined & \textbf{19.37}  & \textbf{9.05} \\
\hlineB{2.5}
\end{tabularx}
\end{center}
\vspace{-1em}
\caption{Ablation Study on SceneEgo test dataset~\cite{wang2023scene}. 
$^\dagger$ denotes the temporal-based method.
\vspace{-0.5em}
}
\label{table:ablation_study}
\end{table}

\noindent\textbf{EgoWholeBody Dataset.}
Compared to existing egocentric datasets, our EgoWholeBody dataset contains diverse body and hand motions, larger quantity of images, and higher image quality. We show this by training our body pose estimation network without our dataset, using the Mo$^2$Cap$^2$~\cite{DBLP:journals/tvcg/XuCZRFST19} and EgoPW~\cite{Wang_2022_CVPR} training dataset. The results, labeled as "w/o EgoWholeBody" in Table~\ref{table:ablation_study}, show that performance without the EgoWholeBody dataset is inferior to our proposed method. This highlights that training with our EgoWholeBody dataset enhances the performance of the pose estimation method. 
We also compare this result with existing methods on the SceneEgo test set (Table~\ref{table:main}). Trained without EgoWholeBody, our approach still outperforms previous methods, showing the effectiveness of our method.

\noindent\textbf{FisheyeViT and Pose Regressor with Pixel-Aligned 3D Heatmap.}
To assess the individual contributions of FisheyeViT and the pixel-aligned 3D heatmap in our single-frame pose estimation pipeline, we perform experiments to measure their impact on the overall performance. First, we substitute the FisheyeViT module in our single-frame pose estimation method to ViT~\cite{dosovitskiy2020image}. The result is shown in ``w/o FisheyeViT'' in Table~\ref{table:ablation_study} and it is worse than our full method. This demonstrates the effectiveness of FisheyeViT in addressing fisheye distortion and feature extraction. 

Next, we analyze the performance of the single-frame pose estimation network when substituting our pose regressor based on pixel-aligned 3D heatmap with the pose estimation heads of previous works~\cite{DBLP:journals/tvcg/XuCZRFST19, DBLP:conf/iccv/TomePAB19, wang2023scene}. The results of the three experiments, labeled as ``w/ Mo$^2$Cap$^2$ head'', ``w/ $x$R-egopose head'' and ``w/ SceneEgo head'', show a performance drop compared to our full method. This emphasizes the crucial role of the pixel-aligned 3D heatmap in accurately estimating egocentric 3D body joint positions.

\noindent\textbf{Diffusion-based Motion Refinement.}
In this ablation study, we assess the effectiveness of our diffusion-based motion refinement with the following experiments: 

First, we compare the performance of our diffusion-based motion refinement with GlobalEgoMocap~\cite{wang2021estimating} by applying the GlobalEgoMocap optimizer on the single-frame body pose estimation results. The result, labeled as ``w GlobalEgoMocap'' in Table~\ref{table:ablation_study}, indicates that our refinement method outperforms GlobalEgoMocap, demonstrating the effectiveness of our diffusion-based motion refinement. 


Second, we remove the uncertainty-aware guidance in the motion refinement. 
Instead, we use fixed Gaussian denoising steps to refine the motion.
The result ``w/o uncert. guidance'' in Table~3, shows that our uncertainty-aware refinement method performs better. Our approach relies on the uncertainty values for each joint, using low-uncertainty joints to guide the generation of high-uncertainty joints. This helps reduce errors in joint predictions caused by egocentric self-occlusion, leading to improved results.

Third, 
we replace our whole-body motion diffusion model with the separate human body and left/right-hand diffusion models and show the accuracy of refined body and hand motion in ``Only body diffusion'' and ``Only hand diffusion'' in Table~\ref{table:ablation_study}. From the results, we observe improvements in the accuracy of motion refined by our whole-body diffusion method, proving that learning the whole-body motion prior can help both the refinement of the body and hand motion by learning the correlation between them.

\section{Conclusion}
In this work, we have introduced an innovative approach to capture egocentric whole-body human motion. Our method comprises a single-frame-based whole-body pose estimation process, which includes FisheyeViT and pixel-aligned 3D heatmap representations. To enhance the initial whole-body pose estimates, we have integrated an uncertainty-aware diffusion-based motion refinement technique.
Our experimental results demonstrate that both our single-frame method and the temporal-based method surpass all existing state-of-the-art techniques in terms of both quality and accuracy.
Looking ahead, we see the potential for extending the applications of FisheyeViT to other vision tasks involving fisheye cameras. Future work could also involve incorporating facial expressions in whole-body motion capture.

\noindent\textbf{Limitations.} Due to serious self-occlusion issues, our method may still predict poses suffering from physical implausibility. This can be solved by introducing the physics-aware motion diffusion models or motion refinement models, such as PhysDiff~\cite{yuan2023physdiff} and PhysCap~\cite{shimada2020physcap}.


{
    \small
    \bibliographystyle{ieeenat_fullname}
    \bibliography{main}
}

\clearpage
\setcounter{page}{1}
\maketitlesupplementary

\section{Fisheye Camera Model}\label{supp:fisheye_model}

In this section, we describe the projection and re-projection function of Scaramuzza's fisheye camera model~\cite{scaramuzza2006toolbox} as follows:

The projection function $\mathcal{P}(x, y, z)$ of a 3D point $[x, y, z]^T$ in the fisheye camera space into a 2D point $[u, v]^T$ on the fisheye image space can be written as:
\begin{equation}
	[u, v]^T = f(\rho) \frac{[x, y]^T}{\sqrt{x^2 + y^2}} 
\end{equation}
where $\rho = \arctan(z / \sqrt{x^2 + y^2})$ and $f(\rho) = k_0 + k_1 \rho + k_2 \rho^2 + k_3 \rho^3 +\dots$ is a polynomial obtained from camera calibration.

Given a 2D point $[u, v]^T$ on the fisheye images and the distance $d$ between the 3D point $[x, y, z]^T$ and the camera, the position of the 3D point can be obtained with the fisheye reprojection function $\mathcal{P}^{-1}(u, v, d)$: 

\begin{equation}
	[x, y, z]^T = d\frac{[u, v, f'(\rho')]^T}{\sqrt{u^2 + v^2 + (f'(\rho'))^2}}
\end{equation}
where $\rho' = \sqrt{u^2 + v^2}$ and $f'(\rho) = k'_0 + k'_1 \rho + k'_2 \rho^2 + k'_3 \rho^3 +\dots$ is another polynomial obtained from camera calibration.
The calibration of the fisheye camera and more details about the fisheye camera model can be found in Scaramuzza~\etal~\cite{scaramuzza2006toolbox}.

Note that a number of different fisheye camera models exist and our method does not depend on one specific fisheye camera model.

\section{Implementation Details} \label{supp:impl_details}
In this section, we describe the implementation details of our methods. We use NVIDIA RTX8000 GPUs for all experiments.

\subsection{FisheyeViT and Pose Regressor with Pixel-Aligned 3D Heatmap}
\subsubsection{Network Structure}
\paragraph{FisheyeViT}
In FisheyeViT, we first undistort the image patches with the method described in Sec.~\ref{method:fisheyevit}, then put the patches into a ViT transformer. In the ViT transformer, the embedding dimension is 768, the network depth is 12, the attention head number is 12, the expansion ratio of the MLP module is 4, and the drop path rate is 0.3. The output sequence from the transformer (with a length equal to 256) is reshaped to a 2D feature map with size $16\times16$.

\paragraph{Pose Regressor with Pixel-Aligned 3D Heatmap}
In the pixel-aligned heatmap, we first use two deconvolutional modules to up-sample the feature map from the FisheyeViT. The first deconv module contains one deconv layer with 768 input channels and 1024 output channels, one batch normalization layer, and one ReLU activation function. The deconv layer's kernel size is 4, the stride is 2, the padding is 1, and the output padding is 0. The second deconv module contains one deconv layer with 1024 input channels and 15$\times$64 output channels, one batch normalization layer, and one ReLU activation function. The hyper-parameters of the deconv layer in the second module are the same as that in the first one.

These deconvolutional modules converts the features from shape $(C\times N \times N) = (768 \times 16 \times 16)$ to shape $(J\times D_h \times H_h \times W_h) = (15 \times 64 \times 64 \times 64)$. Then the soft-argmax function and fisheye reprojection function are applied to get the final body pose prediction.

\subsubsection{Training Details}

In this section, we introduce the training of our single-frame human body pose estimation network, \ie the FisheyeViT and pose regressor with pixel-aligned 3D heatmap. The ViT network in FisheyeViT is initialized with the training weight from ViTPose~\cite{xu2022vitpose} and the pose regressor is initialized with normal distribution, whose mean is 0 and standard deviation is 1. The network is trained on the combination dataset of EgoWholeBody and EgoPW. The ratio between the EgoWholeBody and EgoPW datasets is 9:1. The network is trained for 10 epochs with a batch size of 128, a learning rate of $1e^{-4}$ with the Adam optimizer. 

\subsection{Hand Detection Network}
As described in \cref{method:hand_detection}, we use our EgoWholeBody dataset for training the ViTPose network to regress the heatmap of 2D hand joints. Based on the 2D hand joint predictions, we get the center $\mathbf{C}_{lh}$, $\mathbf{C}_{rh}$, and the size $d_{lh}$, $d_{rh}$ of the square hand bounding boxes. We use the ViTPose network for the simplicity of implementation. Other detection methods can also be used for training the hand detection network. Taking the left hand as an example, we use the bounding center $\mathbf{C}_{lh}$ as the image patch center in \textbf{Step 1} of FisheyeViT (\cref{method:fisheyevit}) and use the half of the bounding box size $d_{lh} / 2$ as the offset $d$ in \textbf{Step 2}. After obtaining the projected points of bounding box center $\mathbf{P}^c_{lh}$ and the bounding box edge $\mathbf{P}^x_{lh}$ on the tangent plane $\mathbf{T}_{lh}$, we set the $l$ in \textbf{Step 3} as two times of the Euclidean distance between $\mathbf{P}^x_{lh}$ and $\mathbf{P}^c_{lh}$. Following \textbf{Step 4}, we get the undistorted hand image crop of the left hand $\mathbf{I}_{lh}$.

The hand detection network is trained for ten epochs with a batch size of 128 and a learning rate of $1e^{-4}$ with the Adam optimizer. 

\subsection{Hand Pose Estimation Network}

As described in \cref{method:hand_pose}, we train the hand-only Pose2Pose network in Hand4Whole method~\cite{Moon_2022_CVPRW_Hand4Whole} with EgoWholeBody training dataset to regress the 3D hand pose from hand image crops. During training, we only use the ground truth 3D hand joint positions as supervision to fine-tune the Pose2Pose network that has been pretrained on the FreiHAND dataset~\cite{zimmermann2019freihand}. The hand pose estimation network is fine-tuned for ten epochs with a batch size of 128 and an initial learning rate of $1e^{-5}$ with the Adam optimizer. 

\subsection{Diffusion-Based Motion Refinement}
In \cref{method:diffusion}, we use the transformer decoder in EDGE~\cite{tseng2023edge} as our diffusion denoising network. We disable the music condition in EDGE~\cite{tseng2023edge} by replacing the music features with a learnable feature vector that is agnostic to input. Here we describe the training details and the refinement details of our diffusion model.

\subsubsection{Training Details}

In this section, we describe the details of training the DDPM model~\cite{ho2020denoising} for learning the whole-body motion prior. Given a whole-body motion sequence with 196 frames from training datasets (Sec.~\ref{dataset:training}) represented with joint locations of the human body (with shape $15 \times 3)$ and hands (with shape $21\times 3$), we transform all poses to the pelvis-related coordinate system and align them to make the human body poses facing forward, obtaining the aligned whole-body motion sequence $\mathbf{x}$. The motion sequence $\mathbf{x}$ is normalized and sent to the DDPM model for training. During training, we randomly sample a diffusion step $t \in \{0, 1, ..., T-1\}$, and use the diffusion forward process to generate the noisy motion $\mathbf{x}_t$. Here the $T$ is the maximal diffusion step and we set $T$ as 1000. We finally run the denoising network to get the original motion $\hat{\mathbf{x}}$ and compare the reconstructed human motion $\hat{\mathbf{x}}$ and the original human motion $\mathbf{x}_t$ with \cref{eq:diffusion_simple_objective}. The network is trained for thirty epochs with a batch size of 256 and an initial learning rate of $2e^{-4}$ with the Adam optimizer. 

\subsubsection{Refinement Details}
After obtaining the trained diffusion model, we follow \cref{method:uncertainty_aware} to refine the input whole-body motion. Here we describe how to obtain the uncertainty values for each joint in the human body and hands. We smooth the 3D heatmap predictions with Gaussian smoothness. The standard deviation of the Gaussian kernel is 1. Then we get the 3D heatmap values $\mathbf{HM}$ at the predicted joint locations with the bilinear interpolation. The heatmap values $\mathbf{HM}$ are firstly normalized to range $[0, 1]$ by making the maximal value of $\mathbf{HM}$ equal to 1. The uncertainty values $\mathbf{u}$ is obtained with:
\begin{equation} \label{supp:eq:uncertainty}
    \mathbf{u} = 0.05 \times (1 - \mathbf{HM})
\end{equation}
In this case, the maximal uncertainty value is 0.05. This value is empirically defined to limit the effect of the stochastic diffusion process in motion refinement.


\section{Synthetic Dataset Comparisons} \label{supp:dataset_comparison}
Compared to other egocentric motion capture training datasets, the EgoWholeBody dataset offers several notable advantages (also see Table~\ref{supp:table:datasets_comp}):

\textbf{Larger Amount of Frames}: EgoWholeBody contains a substantially larger quantity of frames, providing an extensive and diverse dataset for training.

\textbf{Inclusion of Hand Poses}: Unlike other datasets, EgoWholeBody includes hand motion data, making it suitable for egocentric whole-body motion capture.

\textbf{High Diversity in Motions and Backgrounds}: The dataset captures a wide range of human motions and diverse background settings, reflecting real-world scenarios.

\textbf{Publicly Available Models, Motions, and Backgrounds}: The models, motions, and backgrounds are all publicly available. Additionally, the data generation pipeline will be made public, enabling researchers to reproduce or modify the dataset for various different tasks.

These advantages position EgoWholeBody as a valuable resource for advancing research in egocentric whole-body motion capture.

\begin{table*}
\begin{center}
\small
\begin{tabularx}{0.97\textwidth} { 
   >{\raggedright\arraybackslash\hsize=.7\hsize}X 
   >{\centering\arraybackslash\hsize=.6\hsize}X 
   >{\centering\arraybackslash\hsize=.6\hsize}X
   >{\centering\arraybackslash\hsize=.9\hsize}X
   >{\centering\arraybackslash\hsize=.6\hsize}X
   >{\centering\arraybackslash\hsize=.9\hsize}X}
\hlineB{2.5}
Training Dataset & Motion Diversity & Frame Numbers & Motion Type & Image Quality & Annotation Type \\
\hline
EgoPW~\cite{Wang_2022_CVPR} & low & 318 k  & body motion & real-world & pseudo ground truth\\
ECHP~\cite{liu2023egofish3d} & low & 75 k & body motion & real-world  & pseudo ground truth\\
\hline
Mo$^2$Cap$^2$~\cite{DBLP:journals/tvcg/XuCZRFST19} & middle & 530 k  & body motion & low  & ground truth\\
$x$R-EgoPose~\cite{DBLP:conf/iccv/TomePAB19} & middle & 380 k & body motion & \textbf{realistic}  & ground truth\\
EgoGTA~\cite{wang2023scene} & low & 320 k  & body motion & low  & ground truth\\
\hline
EgoWholeBody & \textbf{high} & \textbf{870 k}  & \textbf{body + hands motion} & \textbf{realistic} & ground truth\\
\hlineB{2.5}
\end{tabularx}
\end{center}
\caption{Comparison between different training datasets for egocentric body pose estimation.
}
\label{supp:table:datasets_comp}
\end{table*}

\begin{figure}
\centering
\includegraphics[width=1\linewidth]{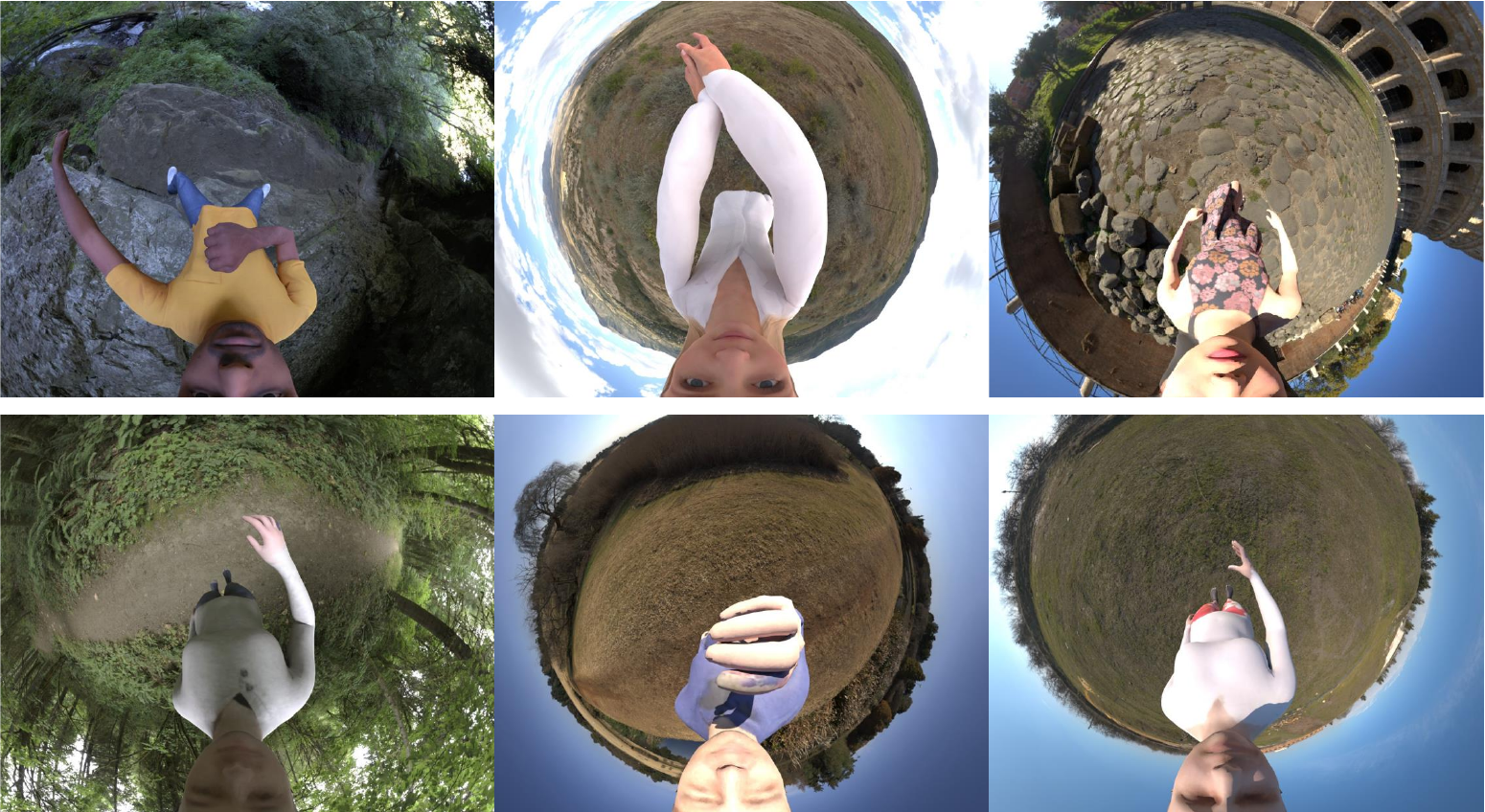}
\caption{Examples of our synthetic dataset EgoWholeMocap. The upper row shows the data rendered with Renderpeople models~\cite{renderpeople}, the lower row shows the data rendered with SMPL-X models~\cite{pavlakos2019expressive}.
\vspace{-1em}
}
\label{supp:fig:synthetic_dataset}
\end{figure}

To show the quality of our synthetic dataset, we also visualize some examples of our synthetic EgoWholeMocap dataset in Fig.~\ref{supp:fig:synthetic_dataset}.

\section{Details of Evaluation Metrics} \label{supp:evaluation_metrics}
In this section, we give a detailed explanation of the evaluation metrics used in our method. Mean Per Joint Position Error (MPJPE) is the mean of Euclidean distances for each joint in the predicted and ground truth poses. 

For the Mean Per Joint Position Error with Procrustes Analysis (PA-MPJPE), we rigidly align the estimated pose to the ground truth pose with Procrustes analysis~\cite{kendall1989survey} and then calculate MPJPE. 

We also evaluate the BA-MPJPE, i.e. the MPJPE with aligned bone length. For BA-MPJPE, we first resize the bone length of predicted poses and ground truth poses to the bone length of a standard human skeleton. Then, we calculate the PA-MPJPE between the two resulting poses. 

\section{Details of Evaluation Datasets} \label{supp:evaluation_dataset}
In our experiment in Sec.~\ref{experiments:main_comparison}, we use three evaluation datasets including SceneEgo test dataset~\cite{wang2023scene}, GlobalEgoMocap test dataset~\cite{wang2021estimating} and Mo$^2$Cap$^2$ test dataset~\cite{DBLP:journals/tvcg/XuCZRFST19}. 

The SceneEgo test dataset contains around $28$K frames of 2 persons performing various motions such as sitting, walking, exercising, reading a newspaper, and using a computer. This dataset provides ground truth egocentric camera pose so that we can evaluate MPJPE on it. This dataset is evenly split into training and testing splits. We finetuned our method on the training split before the evaluation. 

The GlobalEgoMocap test dataset~\cite{wang2021estimating} contains $12$K frames of two people captured in the studio. The Mo$^2$Cap$^2$ test dataset~\cite{DBLP:journals/tvcg/XuCZRFST19} contains $2.7$K frames of two people captured in indoor and outdoor scenes. These two datasets do not provide ground truth egocentric camera poses, thus we first rigidly align the predicted body poses and ground truth body poses and then evaluate PA-MPJPE and BA-MPJPE.

\section{Full Comparison with Existing Egocentric Pose Estimation Methods} \label{supp:full_comparison}

In \cref{experiments:main_comparison}, we mention that we re-train the previous methods with our training data. The results are shown in \cref{table:full_main}. In this experiment, since the GlobalEgoMocap~\cite{wang2021estimating} can be applied to refine the egocentric human body motion predicted from any egocentric pose estimation method, we base the method on Mo$^2$Cap$^2$~\cite{DBLP:journals/tvcg/XuCZRFST19} following the original setting in GlobalEgoMocap~\cite{wang2021estimating}. We also do not show the GlobalEgoMocap results in Mo$^2$Cap$^2$ test dataset~\cite{DBLP:journals/tvcg/XuCZRFST19} since it does not provide egocentric camera poses for all of the sequences. Note that our EgoWholeBody dataset does not contain ground truth scene geometry annotations, therefore we freeze the weights of the depth estimation module in SceneEgo~\cite{wang2023scene} and only train the human pose estimation part.

From the results in \cref{table:full_main}, we can show our single-frame method and our refinement method consistently outperforms all of the previous methods, even if they are trained on our new dataset, which further strengthens the claim in our experiment section (\cref{experiments:main_comparison}). 

\begin{table}
\begin{center}
\small
\begin{tabularx}{0.47\textwidth} { 
   >{\raggedright\arraybackslash}X 
   >{\centering\arraybackslash}X 
   >{\centering\arraybackslash}X  }
\hlineB{2.5}
Method & MPJPE & PA-MPJPE \\
\hline
\multicolumn{2}{l}{\textbf{SceneEgo test dataset~\cite{wang2023scene}}} \\
\hline
Mo$^2$Cap$^2$~\cite{DBLP:journals/tvcg/XuCZRFST19} & 200.3 & 121.2 \\
GlobalEgoMocap$^\dagger$~\cite{wang2021estimating} & 183.0 & 106.2 \\
$x$R-egopose~\cite{DBLP:conf/iccv/TomePAB19} & 241.3 & 133.9 \\
EgoPW~\cite{Wang_2022_CVPR} & 189.6 & 105.3 \\
SceneEgo~\cite{wang2023scene} & 118.5 & 92.75 \\
\hline
Mo$^2$Cap$^2$*~\cite{DBLP:journals/tvcg/XuCZRFST19} & 92.20 & 66.01 \\
GlobalEgoMocap*$^\dagger$~\cite{wang2021estimating} & 89.35 & 63.03 \\
$x$R-egopose*~\cite{DBLP:conf/iccv/TomePAB19} & 121.5 & 98.84 \\
EgoPW*~\cite{Wang_2022_CVPR} & 90.96 & 64.33 \\
SceneEgo*~\cite{wang2023scene} & 89.06 & 70.10 \\
\hline
Ours-Single & \underline{64.19} & \underline{50.06} \\
Ours-Refined$^\dagger$ & \textbf{57.59} & \textbf{46.55} \\
\hlineB{2.5}
Method & PA-MPJPE & BA-MPJPE \\
\hline
\multicolumn{2}{l}{\textbf{GlobalEgoMocap test dataset~\cite{wang2021estimating}}} \\
\hline
Mo$^2$Cap$^2$~\cite{DBLP:journals/tvcg/XuCZRFST19} & 102.3 & 74.46 \\
$x$R-egopose~\cite{DBLP:conf/iccv/TomePAB19} & 112.0 & 87.20 \\
GlobalEgoMocap$^\dagger$\cite{wang2021estimating} & 82.06 & 62.07 \\
EgoPW~\cite{Wang_2022_CVPR} & 81.71 & 64.87 \\
EgoHMR~\cite{liu2023egohmr} & 85.80 & -- \\
SceneEgo~\cite{wang2023scene} & 76.50 & 61.92 \\
\hline
Mo$^2$Cap$^2$*~\cite{DBLP:journals/tvcg/XuCZRFST19} & 78.39 & 63.48 \\
GlobalEgoMocap*$^\dagger$~\cite{wang2021estimating} & 75.62 & 61.06 \\
$x$R-egopose*~\cite{DBLP:conf/iccv/TomePAB19} & 106.3 & 79.56 \\
EgoPW*~\cite{Wang_2022_CVPR} & 77.95 & 62.36 \\
SceneEgo*~\cite{wang2023scene} & 76.51 & 61.74 \\
\hline
Ours-Single & \underline{68.59} & \underline{55.92} \\
Ours-Refined$^\dagger$ & \textbf{65.83} & \textbf{53.47} \\
\hlineB{2.5}
\multicolumn{2}{l}{\textbf{Mo$^2$Cap$^2$ test dataset~\cite{DBLP:journals/tvcg/XuCZRFST19}}} \\
\hline
Mo$^2$Cap$^2$~\cite{DBLP:journals/tvcg/XuCZRFST19} & 91.16 & 70.75 \\
$x$R-egopose~\cite{DBLP:conf/iccv/TomePAB19} & 86.85 & 66.54 \\
EgoPW~\cite{Wang_2022_CVPR} & 83.17 & 64.33 \\
Ego-STAN$^\dagger$~\cite{park2023domain} & 102.4 & -- \\
SceneEgo~\cite{wang2023scene} & 79.65 & 62.82 \\
\hline
Mo$^2$Cap$^2$*~\cite{DBLP:journals/tvcg/XuCZRFST19} & 79.76 & 63.53 \\
$x$R-egopose*~\cite{DBLP:conf/iccv/TomePAB19} & 84.92 & 65.39 \\
EgoPW*~\cite{Wang_2022_CVPR} & 78.01 & 62.37 \\
SceneEgo*~\cite{wang2023scene} & 79.32 & 62.77 \\
\hline
Ours-Single & \underline{74.66} & \underline{59.26} \\
Ours-Refined$^\dagger$ & \textbf{72.63} & \textbf{57.12} \\
\hlineB{2.5}
\end{tabularx}
\end{center}
\vspace{-1em}
\caption{
Performance of our method on four different test datasets. Our method outperforms all previous state-of-the-art methods. $*$ denotes the method trained with the datasets in Sec.~\ref{dataset:training}. $^\dagger$ denotes the temporal-based methods. 
\vspace{-0.5em}
}
\label{table:full_main}
\end{table}

\section{The Standard Deviation of Refinement Method} \label{supp:std}

As described in \cref{experiments:main_comparison}, we generate five samples and calculate the mean and standard deviations of the MPJPE values. The results are shown in \cref{table:std}. From the results, we can see the standard deviations of our results are all around 0.003 mm, which is quite small. 
We suppose that the standard deviations of our results are small for two reasons: 

First, our diffusion process is guided by the low-uncertainty joints. The low-uncertainty joints are more likely to follow the initial motion estimations $\mathbf{x}_e$ and guide the diffusion denoising process of other joints to obtain similar values. 

Second, according to \cref{supp:eq:uncertainty}, the maximal uncertainty value is 0.05 (the actual uncertainty value can be even smaller), which means that when $k=0.1$ in \cref{eq:uncertainty_weight}, the $\mathbf{w} \sim 1$ when $t = 100$ for all joints:
\begin{equation}
    \mathbf{w} = 1 / \left(1 + e^{-0.1(100-1000 \times 0.05)}\right) = 0.9933
\end{equation}
This shows that when $t$ is large enough, the denoising process is always initialized by the estimated motion $\mathbf{x}_e$ and the refinement starts when $t < 100$. When $t < 100$, the Gaussian noise added in \cref{eq:diffusion_refinement} is relatively small.
This also means that we can start from diffusion step $t = 200$ for accelerating the diffusion refinement steps.

\begin{table}
\begin{center}
\small
\begin{tabularx}{0.47\textwidth} { 
   >{\raggedright\arraybackslash\hsize=.8\hsize}X 
   >{\centering\arraybackslash\hsize=.6\hsize}X 
   >{\centering\arraybackslash\hsize=.6\hsize}X  }
\hlineB{2.5}
Dataset & MPJPE & PA-MPJPE \\
\hline
SceneEgo-Body & 57.59$\pm$0.003 & 46.55$\pm$0.003 \\
SceneEgo-Hands & 19.37$\pm$0.002 & 9.05$\pm$0.002 \\
\hlineB{2.5}
Dataset & PA-MPJPE & BA-MPJPE \\
GlobalEgoMocap & 65.83$\pm$0.003 & 53.47$\pm$0.002 \\
Mo$^2$Cap$^2$ & 72.63$\pm$0.003 & 57.12$\pm$0.003 \\
\hlineB{2.5}
\end{tabularx}
\end{center}
\caption{The mean and standard deviations of our refinement method. ``SceneEgo-Body'' and ``SceneEgo-Hands'' show the body and hand results on the SceneEgo dataset. ``GlobalEgoMocap'' and ``Mo$^2$Cap$^2$'' shows the human body results on the GlobalEgoMocap and Mo$^2$Cap$^2$ datasets.
\vspace{-1em}
}
\label{table:std}
\end{table}

\section{Different Parameters in Weight Function} \label{supp_mat:weight_function}

\begin{figure}
\centering
\includegraphics[width=1\linewidth]{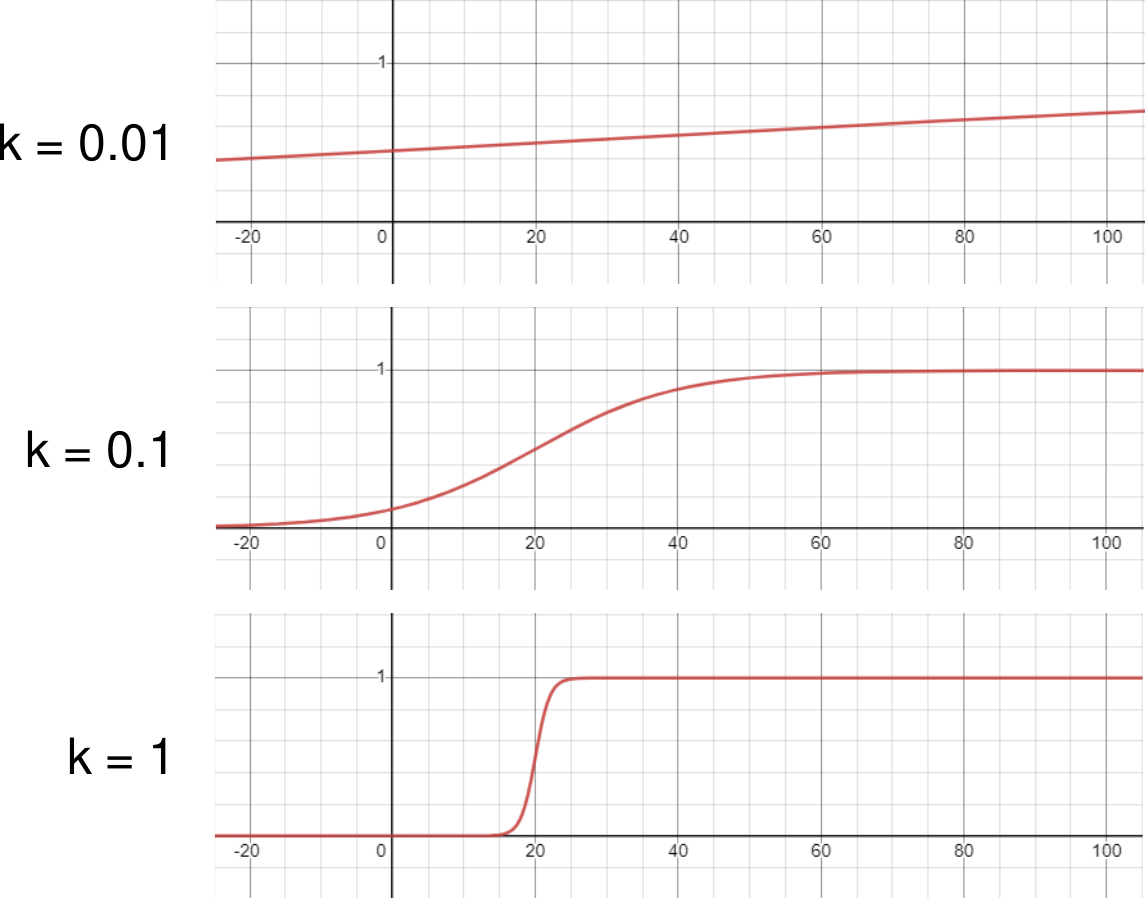}
\caption{The weight function with different hyper-parameters k. The x-axis is the diffusion time step $t$ and the y-axis is the weight $\mathbf{w}$.
\vspace{-1em}
}
\label{supp:fig:different_k}
\end{figure}

In this section, we analyze the effectiveness of parameter $k$ in the weight function~\cref{eq:uncertainty_weight}. We suppose that the uncertainty value of one specific joint is 0.02, then we draw the $\mathbf{w} \text{-} t $ figure in \cref{supp:fig:different_k}. We can observe that when $t \rightarrow 0$, the weight $\mathbf{w}$ is still large when $k = 0.01$. In this case, the initial pose predictions $\mathbf{x}_e$ will significantly affect the final refinement result. When the $k=1$, the weight $\mathbf{w} \sim 0$ when $t < 15$, which makes the diffusion model generate freely without any guidance of the initial joint estimations. This will make the refined motion largely deviate from the initial joint estimations. In our method, we choose a moderate $k = 0.1$, such that the diffusion refinement process can be initially guided by the whole-body pose estimations $\mathbf{x}_e$ and finally refined through the generation of diffusion denoising process.

We also show the results under different $k$ values in~\cref{supp:table:comp_different_k}. The results show that the accuracy of human body poses is the best when $k=0.1$. We also observe that the standard deviations become larger when $k$ is larger. This also demonstrates the above analysis.

\begin{table}
\begin{center}
\small
\begin{tabularx}{0.47\textwidth} { 
   >{\raggedright\arraybackslash}X 
   >{\centering\arraybackslash\hsize=.5\hsize}X 
   >{\centering\arraybackslash\hsize=.5\hsize}X  }
\hlineB{2.5}
Method & MPJPE & PA-MPJPE \\
\hline
k=0.01 & 58.41$\pm$0.001 & 46.92$\pm$0.001 \\
k=0.1 & \textbf{57.59$\pm$0.003} & \textbf{46.55$\pm$0.003} \\
k=1 & 59.90$\pm$0.006 & 48.57$\pm$0.006 \\
\hlineB{2.5}
\end{tabularx}
\end{center}
\caption{Comparison with Spherenet and Panoformer.
\vspace{-1em}
}
\label{supp:table:comp_different_k}
\end{table}



\section{Comparision with networks for panorama images}

In this section, we compare our FisheyeViT network with two other methods dealing with camera distortions, the SphereNet~\cite{coors2018spherenet} and the OmniFusion~\cite{li2022omnifusion}. However, these two methods are designed for panorama images, aiming at the detection task~\cite{coors2018spherenet} and depth estimation task~\cite{li2022omnifusion}. In this experiment, we replace our FisheyeViT with the SphereNet and OmniFusion networks. In SphereNet, we limit the sampling range to the semi-sphere. In OmniFusion, we use the output of the transformer network as the image features and put the image features into our pose regressor. We evaluate the accuracy of the estimated human body pose on the SceneEgo dataset. The results are shown in Table~\ref{supp:table:comp_panorama}, which demonstrates that our FisheyeViT performs better than the previous methods for the distorted images. This might caused by the different patch sampling strategy: our method samples the image patches on the fisheye image $uv$ space, while previous methods samples the patches on the $r\theta\phi$ sphere coordinate system. Our method can generate patches that align well with the layout of egocentric fisheye images and match the design of our pixel-aligned 3D heatmap as mentioned in the introduction: ``the voxels in the 3D heatmap directly correspond to pixels in 2D features, subsequently linking to image patches in FisheyeViT''. However, sampling in the $r\theta\phi$ sphere coordinate system will cause discontinuity due to the \emph{coordinate singularity} of the sphere coordinate system. For example, the neighboring pixels on the fisheye image can be assigned to two patches far away from each other.

\begin{table}
\begin{center}
\small
\begin{tabularx}{0.47\textwidth} { 
   >{\raggedright\arraybackslash}X 
   >{\centering\arraybackslash\hsize=.5\hsize}X 
   >{\centering\arraybackslash\hsize=.5\hsize}X  }
\hlineB{2.5}
Method & MPJPE & PA-MPJPE \\
\hline
SphereNet~\cite{coors2018spherenet} & 90.72 & 75.07 \\
OmniFusion~\cite{li2022omnifusion} & 86.58 & 70.69 \\
\hline
Ours-Single & \textbf{64.19} & \textbf{50.06} \\
\hlineB{2.5}
\end{tabularx}
\end{center}
\caption{Comparison with Spherenet and Panoformer.
\vspace{-1em}
}
\label{supp:table:comp_panorama}
\end{table}

\section{Replacing the Pixel-Aligned 3D Heatmap to MLP}

In this section, we replace our pose regressor with the pixel-aligned 3D heatmap with a simple MLP network. The features extracted with FisheyeViT, with shape $(768 \times 16 \times 16)$ are firstly flattened and we further use two MLP layers to regress the 3D human body poses. The first layer contains one fully connected layer with an output dimension of 1024, one batch normalization layer, and one ReLU activation layer. The second layer contains one fully connected layer with an output dimension of $15 \times 3$. The MPJPE and the PA-MPJPE on the SceneEgo dataset are 130.7 mm and 73.91 mm respectively. This demonstrates the effectiveness of our egocentric pose regressor with pixel-aligned 3D heatmap. 

\section{Compare with Gaussian Smooth}

In this section, we compare our diffusion-based motion refinement method with the simple Gaussian smoothness. The MPJPE and the PA-MPJPE on the SceneEgo dataset are 62.68 mm and 48.87 mm respectively. This demonstrates that our refinement method performs better than the Gaussian smooth approach. This shows that our method relies on motion priors to guide the refinement of human motion, making it more effective than the simple smoothing techniques.

\section{Egocentric Camera Setup}
\begin{figure}
\centering
\includegraphics[width=0.95\linewidth]{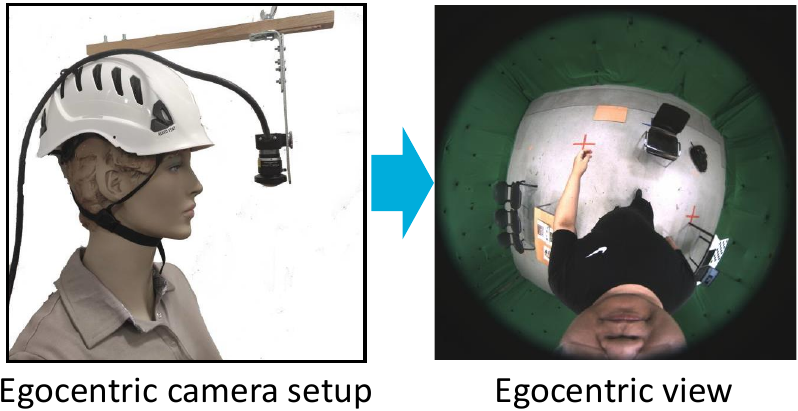}
\caption{The setup of the egocentric fisheye camera and one example of the egocentric image.
\vspace{-1em}
}
\label{supp:fig:setup}
\end{figure}

We use the same egocentric camera setup as previous methods~\cite{DBLP:journals/tvcg/XuCZRFST19, wang2021estimating, wang2023scene, Wang_2022_CVPR}. In this setup, one down-facing PointGrey fisheye camera is mounted in front of the head. 
The illustration is shown in \cref{supp:fig:setup}.

\section{More Visualization Results}

\begin{figure*}
\centering
\includegraphics[width=1\linewidth]{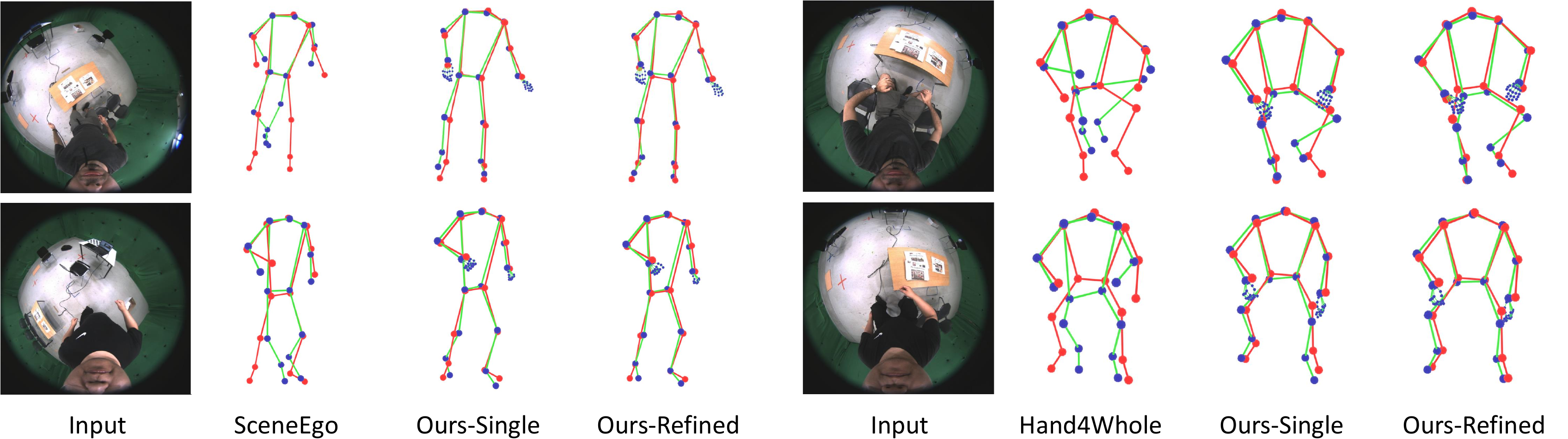}
\caption{Qualitative comparison on human body pose estimations between our methods and the state-of-the-art SceneEgo~\cite{wang2023scene} method. The red skeleton is the ground truth while the green skeleton is the predicted pose. Our methods predict more accurate body poses.
\vspace{-1em}
}
\label{supp:fig:body_results}
\end{figure*}

\begin{figure*}
\centering
\includegraphics[width=1\linewidth]{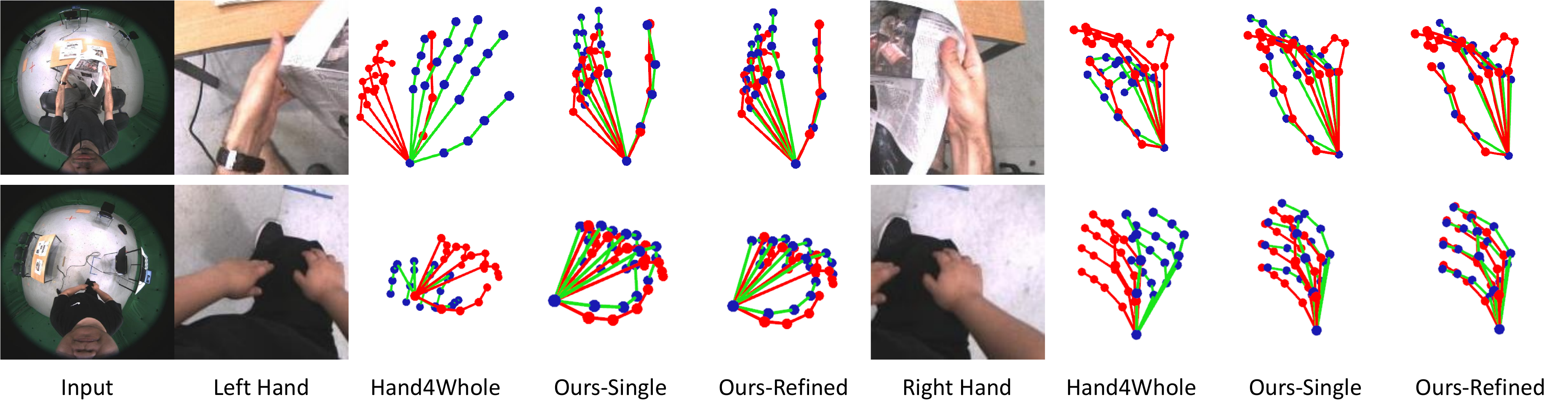}
\caption{Qualitative comparison on hand pose estimation results. Our single-view and refined hand poses are more accurate than the poses from Hand4Whole~\cite{Moon_2022_CVPRW_Hand4Whole} method. The red skeleton is the ground truth while the green skeleton is the predicted pose. 
\vspace{-1em}
}
\label{supp:fig:hand_results}
\end{figure*}

Here we show more results of our methods in \cref{supp:fig:body_results} and \cref{supp:fig:hand_results}.


\end{document}